\documentclass{article} % For LaTeX2e
\usepackage{iclr2018_workshop,times}
\usepackage{hyperref}
\usepackage{url}
\usepackage{amsmath,color}
\usepackage{amssymb}
\usepackage{caption} 
\usepackage{enumerate}% 
\usepackage{graphicx}
\usepackage{upquote}
\usepackage{booktabs} % for professional tables

\title{Visual Analogies between Atari Games for Studying Transfer Learning in RL}

\author{Doron Sobol$^1$, Lior Wolf$^{1,2}$ \& Yaniv Taigman$^2$\\
$^1$ School of Computer Science, Tel-Aviv University\\
$^2$ Facebook AI Research  \\
}

\begin{document}

\maketitle

\begin{abstract}
In this work, we ask the following question: Can visual analogies, learned in an unsupervised way, be used in order to transfer knowledge between pairs of games and even play one game using an agent trained for another game? We attempt to answer this research question by creating visual analogies between a pair of games: a source game and a target game. For example, given a video frame in the target game, we map it to an analogous state in the source game and then attempt to play using a trained policy learned for the source game. We demonstrate convincing visual mapping between four pairs of games (eight mappings), which are used to evaluate three transfer learning approaches. 
%The code and models are available
\end{abstract}

%-------------------------------------------------------------------------
\section{Introduction}
One of the most fascinating capabilities of humans is the ability to generalize between related but vastly different tasks. A surfer will be able to ride a snowboard after much less training than a beginner in board sports; a gamer experienced with adventure games will solve escape rooms long before the one hour is up; and a veteran tennis player will often top the office's ping pong league. The goal of this work is to check if an agent trained via Reinforcement Learning (RL) can gain such an ability using visual analogies: an actor is trained and evaluated on a target task after learning a source task in the typical reinforcement learning setting and is also provided with mappers that given a frame in either game, are able to generate the analogous frame in the other game.

The bidirectional mappers between the video sequences are based on recent computer vision approaches for the task of finding visual analogies, in combination with an added regularization term. We evaluate our methods on two groups of games, and are able to successfully learn the mappers between all same-group pairs. 

Building on the existence of these mappers, we propose several Transfer Learning (TL) techniques for utilizing information from the source game when playing the target game. These methods include techniques such as data-transfer and distillation. Unfortunately, none of these methods seem to be helpful, maybe with the one exception of a method, which trains on scenes that are visually adapted from the source game to the domain of the target game.

Despite the moderate success, we believe that our work presents value to the community in multiple ways. First, we are successful at the challenging video conversion task, which could benefit future efforts. Second, we devise a few possible TL methods that ``almost work''. Third, a critical view of the practical value of TL in the current RL landscape is seldom heard. Lastly, by sharing our results, code, and models, we hope to help others  in minimizing wasted efforts. 
\subsection{Related Work}
\label{sec:related}

While there has been a few contributions that employ reinforcement learning for computer vision tasks, e.g.,~\cite{DBLP:conf/iccv/CaicedoL15} and works that aim to solve perceptual problems in robotics using reinforcement learning, e.g.,~\cite{kwok2004reinforcement}, we are not aware of a work that employs computer vision in order to promote transfer learning in sequential decision making. 

In both computer vision and RL, Generative Adversarial Networks~\cite{DBLP:conf/nips/GoodfellowPMXWOCB14} (GANs) have created a sizable impact. In computer vision, GANs allow for the generation of realistic images using deep neural networks. In reinforcement learning, GANs have opened new avenues for imitation learning and similar forms of transfer by promoting domain confusion, e.g., making sure that the expert and the imitator are playing in indistinguishable ways. 

\noindent{\bf Generating Visual Analogies}
The field of image to image translation has seen major advancement in the last few years with the presentation of new image to image GAN based translation methods such as DiscoGAN~\cite{kim2017learning}, CycleGAN~\cite{lu2017conditional}, UNIT GAN~\cite{liu2017unsupervised} and DistanceGAN~\cite{benaim2017one}. Those methods use some form of unsupervised heuristic alignment loss, to generate analog images in the target domain using an input image from the source domain. %Those methods though do not assume any restrictions on the created images - for example smoothness over time which is needed in the realm of games, or following certain game rules that are needed in our work.

\noindent{\bf Reinforcement Learning Methods}
Recent work in Reinforcement Learning such as the Asynchronous Advantage Actor-Critic (A3C) algorithm~\citep{mnih2016asynchronous} has made running reinforcement learning algorithms faster by allowing it to run on multiple GPUs in parallel. This is done by calculating the gradients of multiple different training episodes in parallel, and made running large amounts of experiments feasible, and therefore made this work possible. While other reinforcement learning algorithms such as rainbow DQN~\cite{hessel2017rainbow}, Double DQN~\cite{van2016deep} and other extensions of DQN~\cite{mnih2013playing} may be more efficient and achieve better results using less training episodes, they are less practical when running large amounts of experiments due to the fact that they can not be paralleled and therefore take more time in a real world scenario.

\noindent{\bf Transfer Learning in Reinforcement Learning}
Transfer learning in reinforcement learning has been studied in~\cite{rusu2016progressive}, where TL takes place between different Atari games by using the activations of every level of the network in the source domain as an additional input for the next layer in the target domain. This work does not consider the visual similarities between the environments. The results are evaluated using the area under the curve of the mean reward over the training session, which is high even in cases where the training is somewhat more successful at earlier stages but there is no improvement in overall performance. 

Concurrently with our work, ~\cite{gamrian2018transfer} have researched the use of unsupervised image to image translation to overcome visual differences between Atari games. In comparison to our work, they used the same base game as the target and the source game and only slightly changed the visual part of the target game by adding a constant noise to the game frames - for example green lines over the screen or a rectangle in a certain location. While in our work we transfer knowledge between completely different games.

\section{Settings}
\label{settings}
Our objective is to train an agent on the target game $T$ in the most efficient way using knowledge from the source game $S$. In phase one the algorithm has unlimited number of diverse frames from both the target game and the source game, but does not have access to the interactive games environments or to the reward functions of the games environments. In addition the algorithm is not aware of the actions taken during the frame's production. In phase two, the algorithm is given the environment of both games and its goal is to train an agent on the target game with the least amount of training episodes from the target game, but with no limitations on the number of training episodes from the source game. In this work, the data of the first phase, namely frames from $t$ and $s$ is used to train (in an unsupervised way) the visual analogies mappers $G: s \rightarrow t$ and $G^{-1}: t \rightarrow s$ . In phase two we are using those mappers and apply different TL methods from $S$ to $T$ in order to train an agent on the target game $T$ in the most efficient way. 

\noindent{\bf Learning Cross-Domain Video Mapping}
 To create visual analogies between a pair of games, we need a large amount of relatively variant frames from both the source and the target games. We collect those frames offline by using an actor. This actor does not need to be an expert and we do not need to know the action it applies or to imitate it. However, it is required that the states that are generated by this agent are diverse enough and therefore the actor is required to remain in the game for a while on different game scenarios. %Those frames are used to learn, in an unsupervised way, a mapper $G: s \rightarrow t$, between the frames of the source game $s$, and the frames of the target game $t$. We also learn the mapper in the reverse direction  $G^{-1}: t \rightarrow s$.

%$s$ AND $t$ WERE NOT DEFINE, ALSO t and s AND IN GENERAL IT IS WRITTEN POORLY

\begin{figure*}[t]
  \begin{tabular}{ccccc}

  \includegraphics[clip,width=0.172\linewidth]{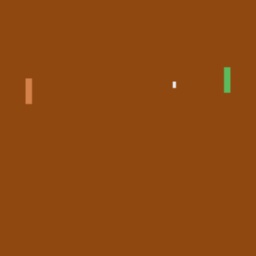}
& 
  \includegraphics[clip,width=0.172\linewidth]{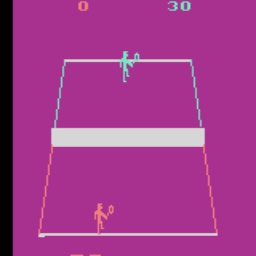}
  &
  \includegraphics[clip,width=0.172\linewidth]{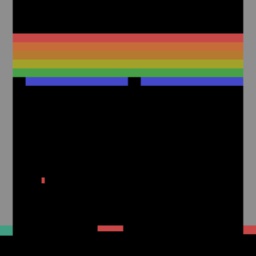}
& 
  \includegraphics[clip,width=0.172\linewidth]{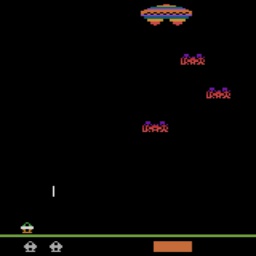}
  &
  \includegraphics[clip,width=0.172\linewidth]{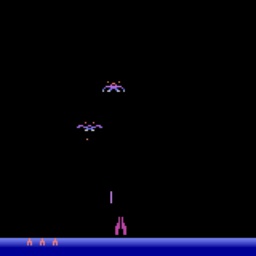}
  \\
% ~~~~~~ (a) & ~~~~ (b) &~~~~~~ (c) & ~~~~ (d)& ~~~~~~ (e) \\
   \includegraphics[clip,width=0.172\linewidth]{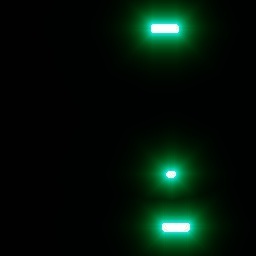}
& 
  \includegraphics[clip,width=0.172\linewidth]{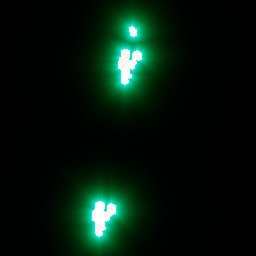}
  &
  \includegraphics[clip,width=0.172\linewidth]{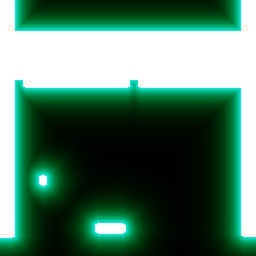}
& 
  \includegraphics[clip,width=0.172\linewidth]{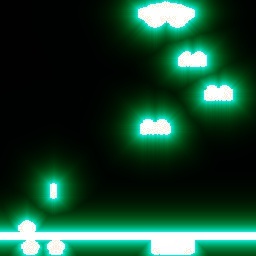}
  &
  \includegraphics[clip,width=0.172\linewidth]{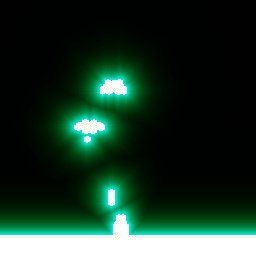}
  \\
  Pong & Tennis& Breakout& Assault& Demon-Attack\\
% ~~~~~~ (f) & ~~~~ (g) &~~~~~~ (h) & ~~~~ (i)& ~~~~~~ (j) \\

  \end{tabular}
  \caption{\label{fig:prep} The obtained attention maps for a frame from each of the five tested games.}
\end{figure*} 

\begin{figure*}

  \begin{tabular}{cc}

  \includegraphics[clip,width=0.485\linewidth]{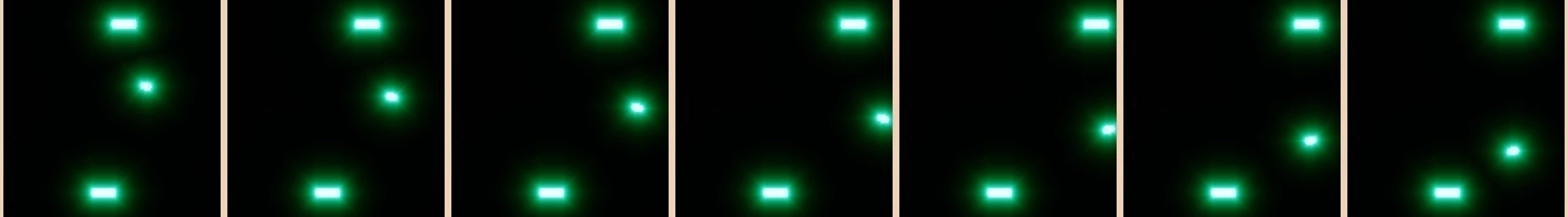}
& 
  \includegraphics[clip,width=0.485\linewidth]{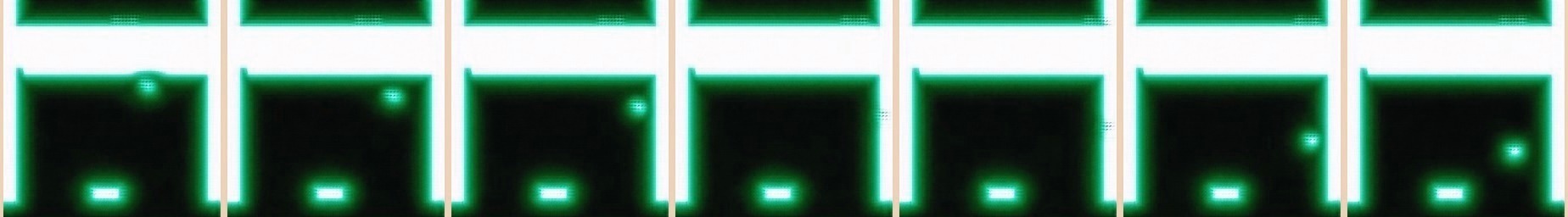}\\
 ~~~~~~ Pong as source & ~~~~ Breakout as target\\
 
  \includegraphics[clip,width=0.485\linewidth]{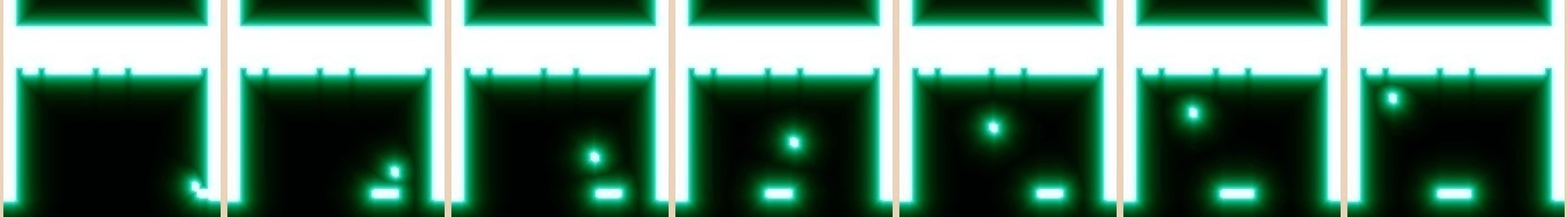}
& 
  \includegraphics[clip,width=0.485\linewidth]{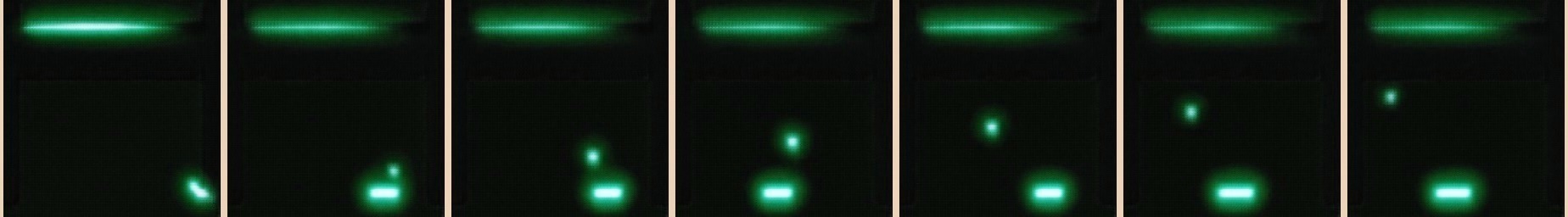}\\
 ~~~~~~  Breakout as source & ~~~~ Pong as target\\
 
  \includegraphics[clip,width=0.485\linewidth]{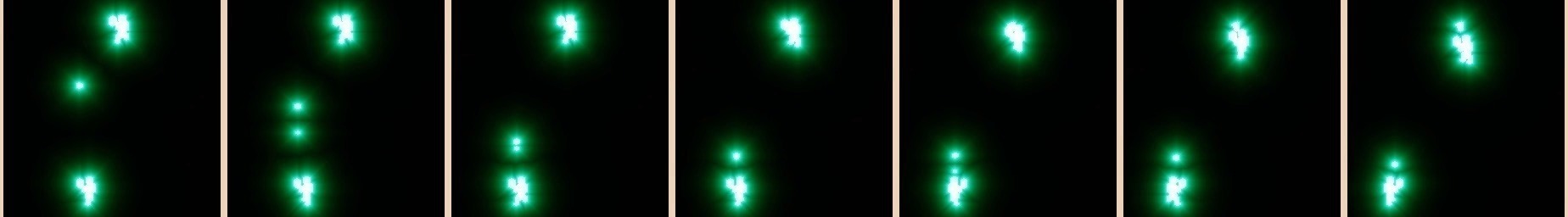}
& 
  \includegraphics[clip,width=0.485\linewidth]{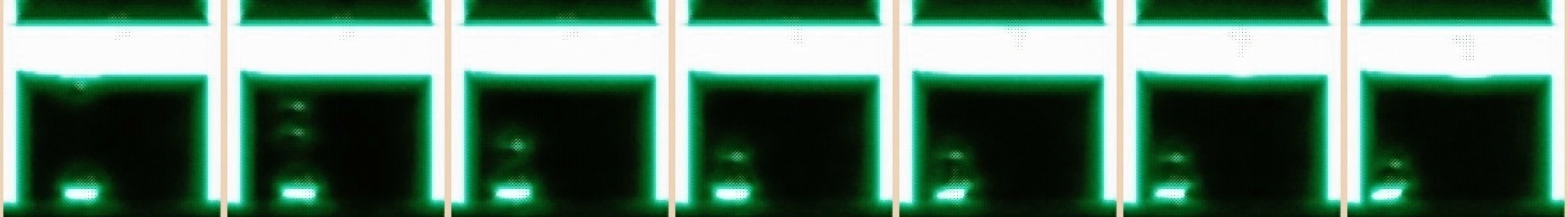}\\
 ~~~~~~ Tennis as source & ~~~~  Breakout as target\\
 
  \includegraphics[clip,width=0.485\linewidth]{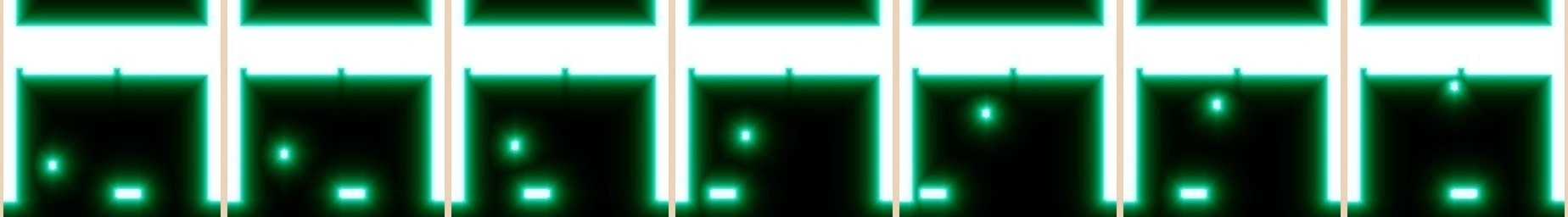}
& 
  \includegraphics[clip,width=0.485\linewidth]{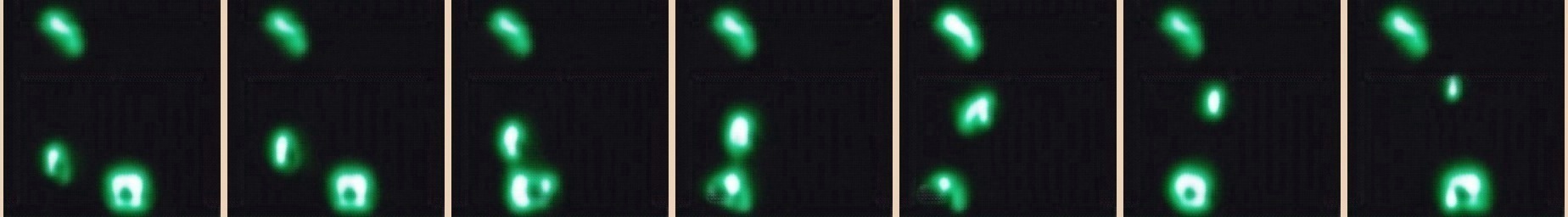}\\
 ~~~~~~ Breakout as source & ~~~~ Tennis as target\\

   \includegraphics[clip,width=0.485\linewidth]{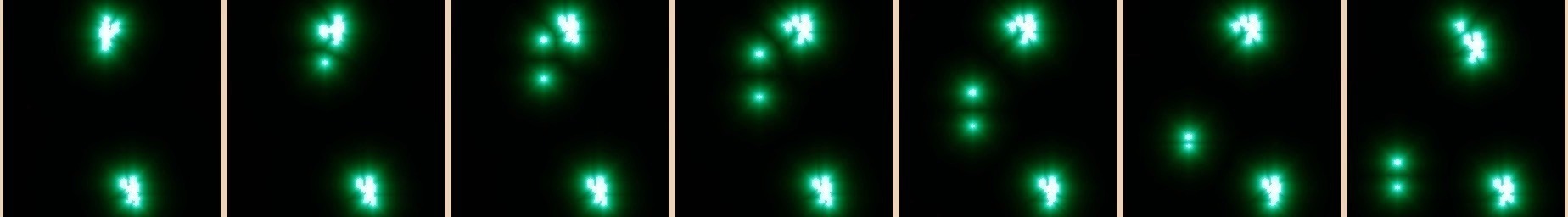}
& 
  \includegraphics[clip,width=0.485\linewidth]{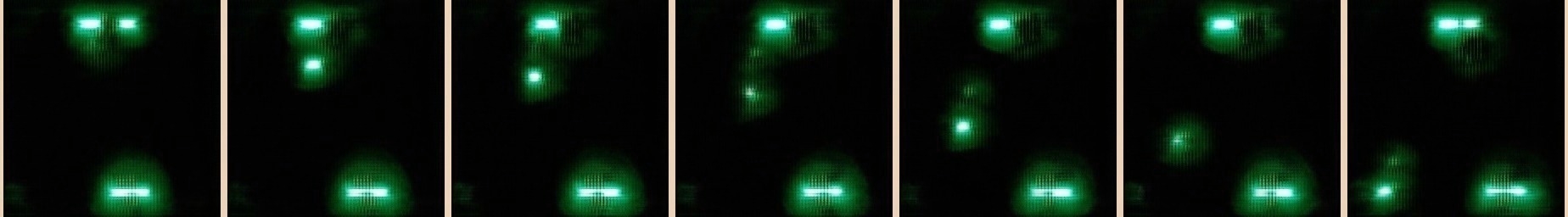}\\
 ~~~~~~ Tennis as source & ~~~~ Pong as target\\
 
  \includegraphics[clip,width=0.485\linewidth]{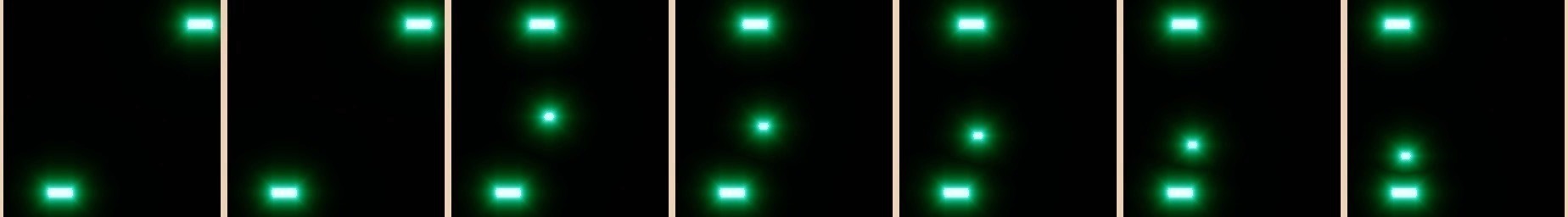}
& 
  \includegraphics[clip,width=0.485\linewidth]{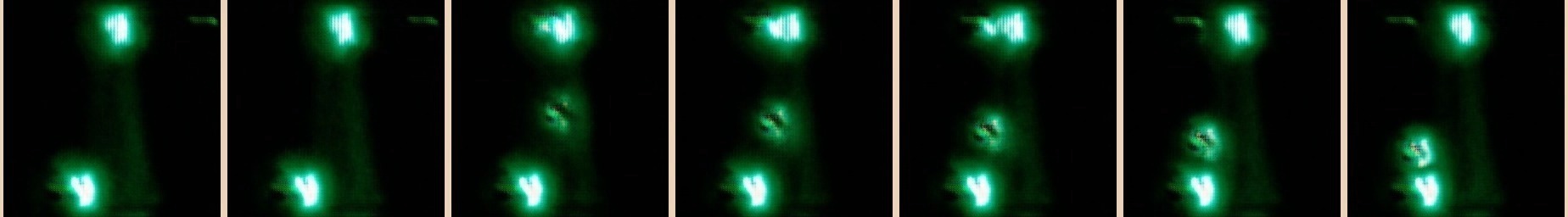}\\
 ~~~~~~ Pong as source & ~~~~ Tennis as target\\

  \includegraphics[clip,width=0.485\linewidth]{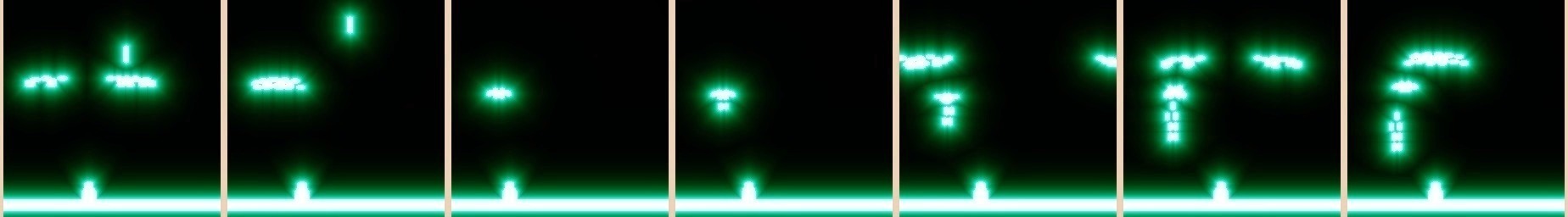}
& 
  \includegraphics[clip,width=0.485\linewidth]{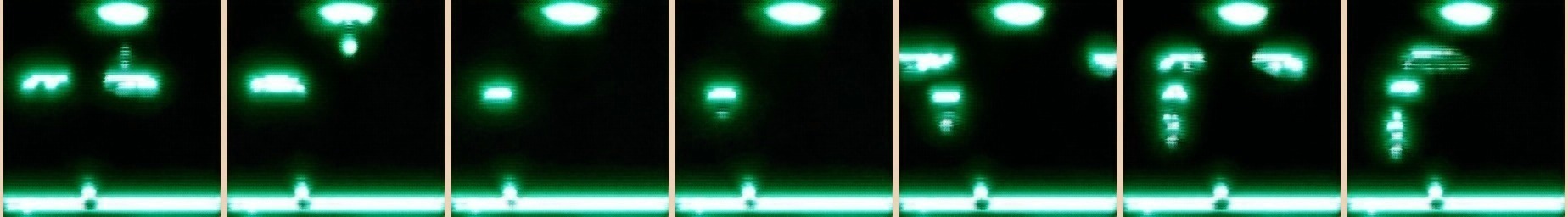}\\
 ~~~~~~ Demon-Attack as source & ~~~~ Assault as target\\
 
   \includegraphics[clip,width=0.485\linewidth]{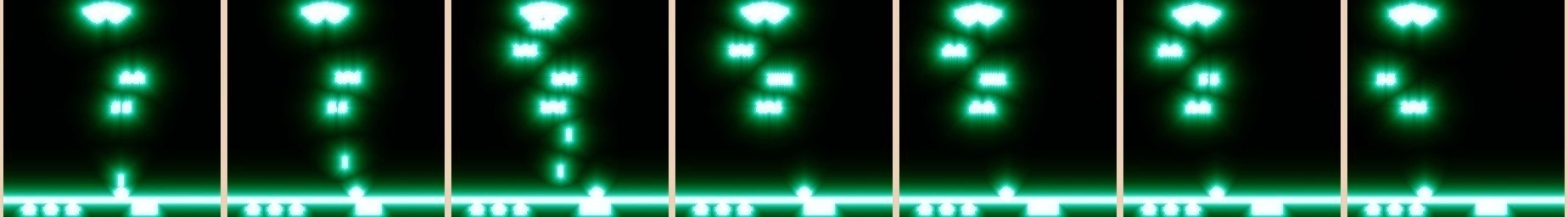}
& 
  \includegraphics[clip,width=0.485\linewidth]{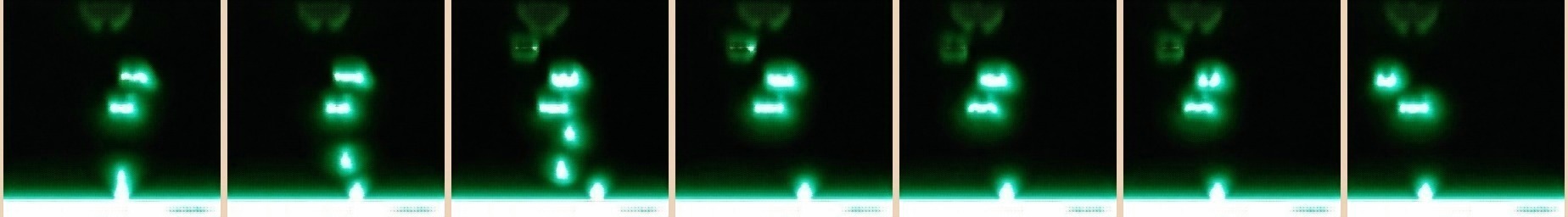}\\
 ~~~~~~ Assault as source & ~~~~ Demon-Attack as target\\
 
  %~~~ \includegraphics[clip,width=0.40\linewidth]%{figures/horse2zebra_distance_full_l1_A_B.png}
%{newfigures/cityscapes_full.png}  & 
 % ~~~ \includegraphics[clip,width=0.40\linewidth]%{figures/horse2zebra_distance_half_l1_A_B.png}\\
  %{newfigures/cityscapes_half.png}\\
 %~~~~~~ (c) & ~~~~ (d)\\
 
  \end{tabular}
  \caption{\label{fig:conv_all} Images of samples of consecutive frames from the source game (left) to the target game (right of the same line).}
\end{figure*}

\label{sec:video}
% In order to generate the mapper functions $G$ and $G^{-1}$ in the unsupervised learning step we use the recent advancement in the unsupervised image to image translation problem and use the UNIT GAN architecture.
% %BUT WE USE UNIT AND NOT CYCLEGAN, WHICH IS BTW, ONE WORD. 
% Using UNIT GAN by itself, with only the cycle consistency loss heuristic, and shared latent space heuristic from the group of frames of $s$ to the group of frames of $t$ does not immediately work and requires some improvements: 
% %THIS PARAGRAPHS IS WRITTEN VERY BADLY

Despite recent advancement in unsupervised mapping between visual domains, the task of mapping between games on the raw image data is still too challenging. We therefore apply the following preprocessing steps, depicted in Fig~\ref{fig:prep}:
%ENGLISH. 
% and is applied as following: 
\begin{enumerate}
\item Rotating the frames, if needed, so that the main axis of motion is horizontal. For example in the game Pong the paddle movement is vertical and so we rotate the frames by 90 degrees in order for it to match the paddles movement of breakout and tennis.
\item  Extracting the important features of the games by applying an attention operator to the frame, this is done by subtracting either the median pixel value at each location or the median pixel value of the entire image (depending on the game), and then applying a threshold to obtain a binary image.
\item Enlarging the relevant features in the frame by applying a dilation filter on the image. This is needed due to the fact that some small items, even 1-2 pixels in size, can be very important (e.g., bullets fired or the ball).
\item Creating three channels with varying visual scales by cloning the dilated image and applying two levels of blurring.
\end{enumerate}

In order to perform the visual mapping, we modify the UNIT GAN method~\cite{liu2017unsupervised}. Namely, we train the mapper functions $G$ and $G^{-1}$, using the network architecture of UNIT GAN.
%WHICH LOSS ARE USED NEED EQUATIONS. 
The UNIT GAN architecture uses two encoders $E_s$ and $E_t$ and two decoders $Dec_s$ and $Dec_t$ such that $Dec_s \circ E_s$ is a variational auto encoder (VAE) of domain $s$ and $Dec_t \circ E_s$ is the mapper $G$. This yields the following losses for each domain , (Those losses are for the direction from $s$ to $t$ but the same losses apply in the opposite direction and trained simultaneously):
\[L_{VAE_t} = KL(p_t(z_t|x_t)||p_{\eta}) - \mathbb{E}_{z_t \sim p_t(z_t|x_t ) }\left[ \log p_{Dec_t} (x_t|z_t) \right] \]
Where $p_t(z_t|x_t )$ is the distribution of the latent vector, the prior distribution, $p_{\eta}$ is a zero mean Gaussian distribution and $p_{Dec_t}$ is the distribution that $Dec_t$ generates.\\
The GAN loss on the images generated from the generators:
\[
	L_{GAN_t} = \mathbb{E}_{x_t \sim P_t} \left[ log D_t(x_t)\right] + \mathbb{E}_{z_{s} \sim p_{s}(z_{s}|x_{s} ) }\left[ 1 - \log D_{t}(Dec_{t}(z_{s}))) \right]
\]
Where $D_t$ is the discriminator of domain $t$.

The cycle consistency loss which means that the latent vector of a sample $x_t \in t$ is the same as that of $Dec_t(E_s(Dec_s(E_t(x_t))))$, is given by:
\begin{align*}
L_{cc_t} = & KL(p_{s}(z_{s}|x_{s})||p_{\eta}) + KL (p_t(z_t|Dec_t(E_{s}(x_{s}) ||p_{\eta})  )) -  
  \\
&\mathbb{E}_{z_t \sim p_t(z_t|Dec_t(E_{s}(x_{s}) ) }\left[ \log p_{Dec_{s}} (x_{s}|z_t) \right]
\end{align*}

Each encoder $E_s/E_t$ contains three encoding layers followed by three residual blocks followed by one shared layer for both encoders. The decoders are a mirror of those layers in the reverse order and use de-convolutions instead of convolutions.

Those losses, by themself, fail to produce a desirable solution due to mode collapse. In order to fix this, we add the gradient-penalty regularization term of improved WGAN~\citep{gulrajani2017improved}, adapted to the problem of cross-domain mapping:
\[ L_{GP} = \mathbb{E}[(||\nabla_{\hat{x}}D(\hat{x})||^2-1)^2]\] 
where 
$\hat{x}$ is either $ \hat{s}=\epsilon s+(1-\epsilon)G^{-1}(t) $ or $ \hat{t}=\epsilon t+(1-\epsilon)G(s) $
, $D$ is the GAN's discriminator, $s$ are samples of $S$, $t$ are samples of $T$, and $\epsilon \sim U[0,1]$.
Therefore we optimize on:
\begin{align*}
 \min\limits_{E_s, Dec_s, E_t, Dec_t} ~ \max\limits_{D_s, D_t} ~ & L_{VAE_s}(E_s, Dec_s) + L_{VAE_t}(E_t, Dec_t) + L_{GAN_s}(E_s, Dec_s, D_s) \\ & + L_{GAN_t}(E_t, Dec_t, D_t) 
  + L_{cc_s}(E_s, Dec_s, E_t, Dec_t) \\ & + L_{cc_t}(E_t, Dec_t, E_s, Dec_s) +  L_{GP_s}(D_s)  + L_{GP_t}(D_t) 
\end{align*}
This regularization term on the discriminator makes the discriminator within the
space of 1-Lipschitz functions, which makes the problem optimized by the GAN to be a Wasserstein distance function which has better theoretical properties, and makes the GAN converge to a better function and not to collapse. The weights we use to balance between the various loss terms are fixed throughout the experiments.

%We notice that without the addition of the improved W-GAN regularization loss, and preprocessing the attention map, the GAN collapses and practically learns nothing.

\section{Transfer Learning Methods}
\label{sec:tlmethods}

Training the strategy $\pi_t$ for the target game and the strategy $\pi_s$ of the source game (when used), is done with the asynchronous advantage actor-critic (A3C) algorithm~\citep{mnih2016asynchronous}. 
%{\color{red}The network architecture consists of four convolutional then max-pool than relu activation layers followed by an LSTM layer with 512 activations and two fully connect layers for the predicted action and value.} 
The network architecture consists of four consecutive layers of convolutions, each layer $i$ with kernels of size $k_i$ (see below), and depth of size $d_i$; each of the layers is followed by max pooling of size two and the ReLU activation function. The convolutional stack is followed by a recurrent LSTM layer with $\lambda$ hidden neurons. The network is topped with two fully connected layers, used to predict the action and its corresponding value. %{\color{red} TODO put $\lambda=512$, $n_1$ and $n_2$ at the experiment section}\\

We evaluated various methods of transferring knowledge between games, and those are the most promising three methods: (i) Data Transfer by Pretraining, which uses pretraining on the converted source game in order to bootstrap the training process, (ii) Continuous Data Transfer, which uses episodes from the converted source game as an additional training data, and (iii) Distilation, which uses the visual mapping obtained by $G^{-1}$ between the source and the target, and $\pi_s$ the policy of the source game, in order to initialize the network activations of the target agent. 

{\bf Data transfer by pretraining}
\label{sec:pretrain}
%Being in the same physical world, the assumption that the source game is a remote approximation of the target game is logical. Such assumption implies that a good agent for the source game will perform reasonably well on the target game, and that pretraining an agent on the source game may help boost its training on the target game at later stages. The following procedure is used for the data transfer: 
In every pair of games we chose, there are major similarities between both the rewards and the game dynamics. Therefore, we can assume that after applying the visual transformation $G$ on the source game $S$, the converted source game is a remote approximation of the target game. This transfer method finetunes, on the target game, an agent that was pretrained of the converted source frames: We first transform frames from the source game $s$ using $G$ to get the remote approximation of the target game. We then train a policy $\pi_t$  using the A3C algorithm on these converted frames using the reward of the source game and using a static mapping of actions from the source game actions to the target game actions. We then fine tune the resulting policy on the target game, obtaining $\pi_t$. By a static mapping of the actions we mean, for example, that UP in the Pong game might correspond to RIGHT in Breakout and therefore the actions need to be adjusted. If the method works, one can try all possible permutations and select the best mapping. In our experiments, we provide this mapping, thus providing an upper bound on the actual performance.

{\bf Continuous data transfer}
\label{sec:contin}
In this method, we do not separate the training into multiple phases and instead provide a continuous training signal from the source game while training $\pi_t$. Namely, we train the target policy $\pi_t$ on mixed samples from both $G(s)$ and $t$ throughout the entire training process. This method could potentially help the agent overcome overfitting by creating a more general agent.

{\bf Distilation}
\label{sec:distilation}
The methods above employed the mapping $G$ and trained in the visual domain of the game $T$. This method employs $G^{-1}$ and a pretrained source game policy $\pi_s$. A direct approach of fine-tunning $\pi_s \circ G^{-1}$, leads to a very large network, mainly since the image to image GAN $G$ is relatively deep and consumes most of the training time, even though it is not being updated. We therefore train a network $\pi_{mimic}$ of the same architecture as $\pi_t$ in order to mimic $\pi_s \circ G^{-1}$, on the set unsupervised frames from the target game. We extract the convolution layers from the mimicking network and use them to initialize the convolution layers of a new network $\pi_t$. We then continue to train $\pi_t$ using game data from the target domain. Therefore we distill the spatial information learned by $\pi_s$ into a new network $\pi_t$, which is then fine-tuned on the target game (using all layers of $\pi_{mimic}$ provides inferior performance).

\section{Results}
\label{sec:experiments}

The experiments are conducted on five Atari games, split into two groups. The first group contains the games Breakout, Tennis and Pong, in which the player has a paddle it controls and its goal is to achieve a certain objective by hitting the ball in a certain way. The second group of games are Demon-Attack and Assault, in which the player has a spaceship it controls and it needs to shoot the targets (similar to Space Invaders). Despite some effort, we were not able to identify other potential pairs among the Atari games for TL. The two groups give rise to four pairs of games, which yield eight transfer directions.

The Experiments are conducted on every pair of games in the following way:
\begin{enumerate}
\item A big number of samples are sampled from both games using the untrained agent described in Sec.~\ref{settings}.
\item We train the mappers $G$ and $G^{-1}$ using the method described in Sec.~\ref{sec:video}. We manually stop the training when we achieve visually adequate results.
%until we get good enough results (checked visually).
\item 
%Using this mapper we use the Asynchronous Advantage Actor Critic algorithm with our transfer learning methods described in Sec.~\ref{sec:tlmethods} to train $\pi_t$. {\color{red} 
We train $\pi_t$ using one of the three transfer learning methods described in Sec.~\ref{sec:tlmethods}.
\end{enumerate}

In Fig.~\ref{fig:conv_all} there are samples of the transferred frames between the target game and the source game and vice versa which were obtained by using the method described in Sec.~\ref{sec:video}, the images show the mapping between sequences in the source game (left) to the converted images of the target game. As can be seen from the correlation between the balls locations, and the paddle or the tennis player in the games Tennis, Pong and Breakout, and from the correlations between the spaceships and the enemies locations in the games Assault and Demon-Attack, the mappings obtained using the improved UNIT GAN seem to convey the semantics of the games and to align frames from the source game to the target game in a reasonable fashion. %, Fig.\ref{fig:conv_all} and seems to be successful.

The architecture of the policy networks $\pi_t$ and $\pi_s$ is described in Sec.~\ref{sec:tlmethods}. We set $\lambda = 512$, $k_1 =k_2= 5$, $k_3 = 4$ and $k_4 = 3$, $d_1 =d_2= 32$, and $d_3 = d_4 = 64$, and max pooling with window size $2$. In the RL phase, the training of the agent is done using the A3C algorithm with 16 processes for training and 6 processes for the test in order to achieve stable results. When using a transfer learning method that uses pretraining on the source domain, the source game is pre trained using the same parameters as the target game until it converges. As for the details of the games environment itself we use the deterministic version of the games, and in order to speed up training we skip some of the frames so that the agent is only looking at every fourth frame. When training with the continuous data transfer method, we train with 12 processes for the converted source and 4 processes for the target. We do that since training on the converted source game requires more processing power as it also needs to transfer the frames, and this division empirically yields the best results. Lastly, when using the distillation transfer method, we start by sampling frames from the target games. We do that using the same untrained agent that was used for sampling the unlabeled frames in Sec.~\ref{sec:video} . Then, for every sample $t \in T$ we calculate $\pi_s(G^{-1}(t))$ as the labels of $t$, and train $\pi_{mimic}$ using cross entropy loss.

\begin{figure*}
\centering
\begin{sc}
  \begin{tabular}{cc}

  \includegraphics[clip,width=0.4\linewidth]{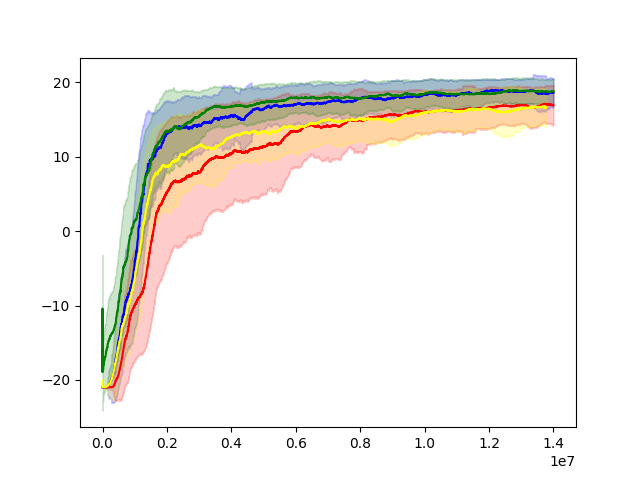}
& 
  \includegraphics[clip,width=0.4\linewidth]{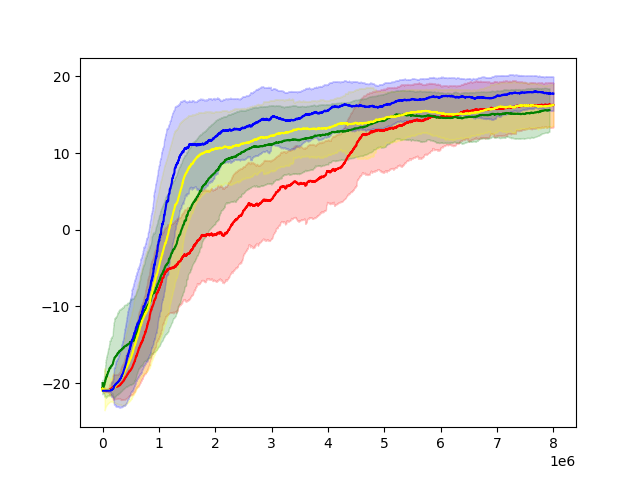}\\
 ~~~~~~ Breakout $\Rightarrow$ Pong & ~~~~ Tennis $\Rightarrow$ Pong \\
 
   \includegraphics[clip,width=0.4\linewidth]{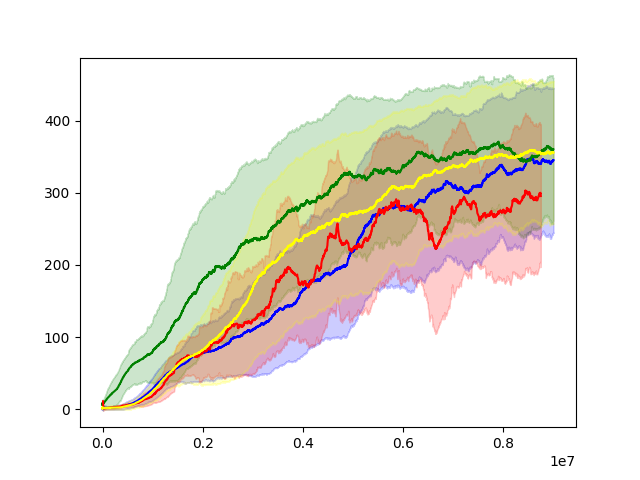}
& 
  \includegraphics[clip,width=0.4\linewidth]{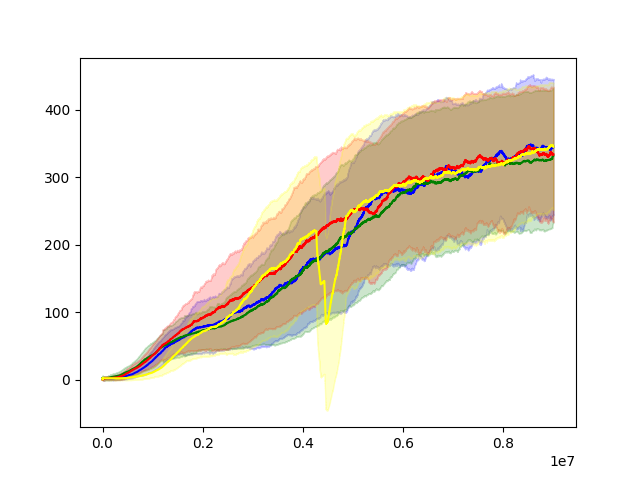}\\
 ~~~~~~ Pong $\Rightarrow$ Breakout & ~~~~ Tennis $\Rightarrow$ Breakout\\
 
   \includegraphics[clip,width=0.4\linewidth]{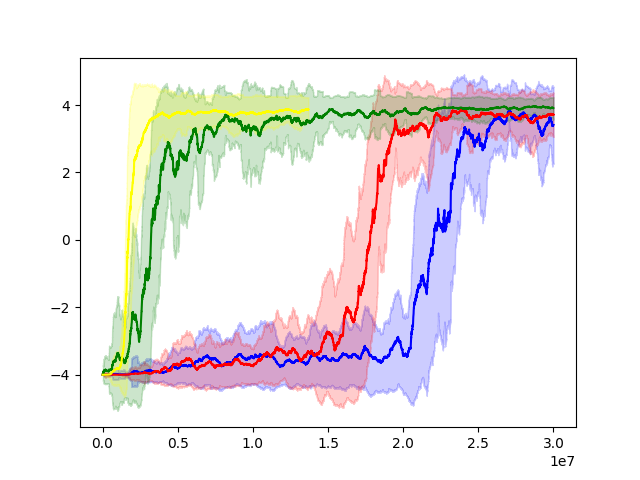}
& 
  \includegraphics[clip,width=0.4\linewidth]{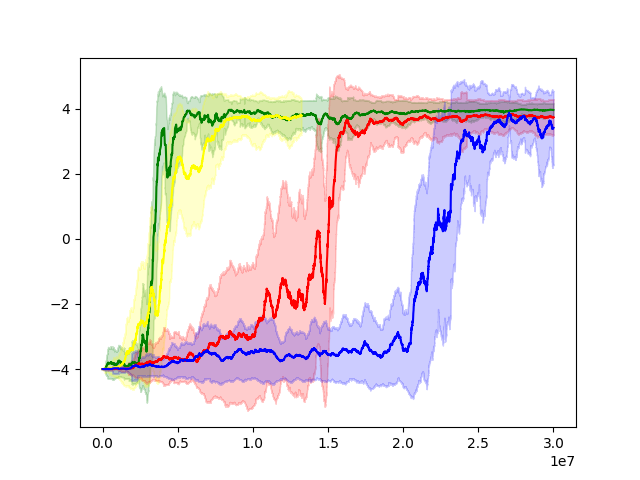}\\
 ~~~~~~ Pong $\Rightarrow$ Tennis & ~~~~ Breakout $\Rightarrow$ Tennis\\
   \includegraphics[clip,width=0.4\linewidth]{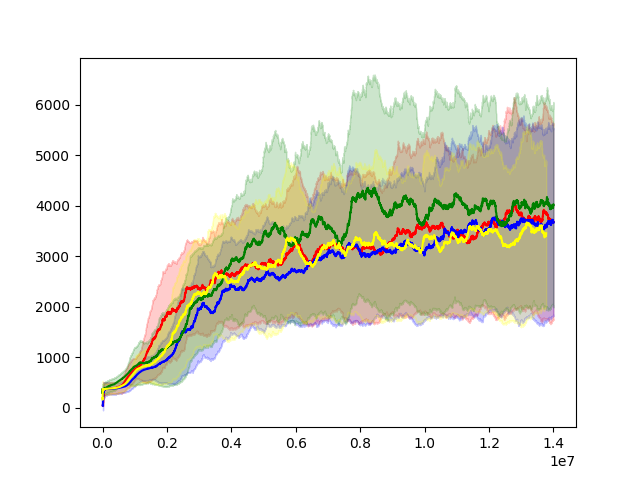}
& 
  \includegraphics[clip,width=0.4\linewidth]{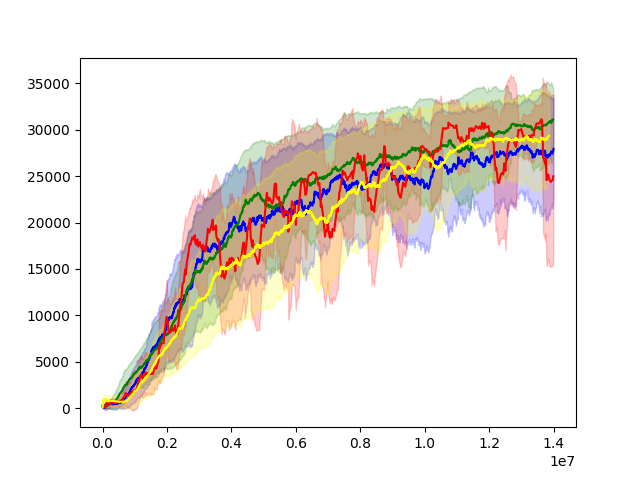}\\
 ~~~~~~ Demon-Attack $\Rightarrow$ Assault & ~~~~ Assault $\Rightarrow$ Demon-Attack\\
 
  %~~~ \includegraphics[clip,width=0.40\linewidth]%{figures/horse2zebra_distance_full_l1_A_B.png}
%{newfigures/cityscapes_full.png}  & 
 % ~~~ \includegraphics[clip,width=0.40\linewidth]%{figures/horse2zebra_distance_half_l1_A_B.png}\\
  %{newfigures/cityscapes_half.png}\\
 %~~~~~~ (c) & ~~~~ (d)\\
 
  \end{tabular}
  \end{sc}
  \caption{\label{fig:graphs} A comparison of the training logs for the various TL methods. The x-axis is the number of training steps, and the y-axis is reward. The plots are averaged over three independent runs where the transparent color is the variance. The blue line is the baseline, the red is distillation, the yellow is continuous data transfer and the green is data transfer for pretraining.}
\end{figure*}

The success of the TL methods is shown in Fig.~\ref{fig:graphs}, which depicts the training progress (local average reward as a function of the number of target domain sessions). Note that when trained using the second method (continues data transfer), we only count the training episodes from the target game. The results are not conclusive and we employ a subjective rating scale for describing the level of success of the TL methods on a given pair of games. A method is considered successful in transferring knowledge from the source game, if reaching a certain level of performance requires less supervised training samples in the target domain than the baseline method of vanilla training in that domain. We distinguish between three levels of success: 
\begin{enumerate}
\item Upon convergence or reaching the maximum possible reward, the method that employs TL outperforms the baseline method - either in the actual reward or in the number of training episodes needed to reach that reward.
\item The TL method achieves almost all levels of performance between the random performance and the converged performance with less samples than the vanilla method.
\item The TL method achieves non-trivial levels of performance faster than the baseline method but then stops leading. 
\end{enumerate}
We also employ a star (*) to denote situations in which the TL method starts off, without any supervised samples from the target domain, in a level that is significantly better than random. This can happen with any level of success. In addition we employ a dash (-) to indicates a lack of success.

\begin{figure*}

  \begin{tabular}{cc}
   \includegraphics[clip,width=0.485\linewidth]{graphs/pong_from_breakout_var.png}
& 
  \includegraphics[clip,width=0.485\linewidth]{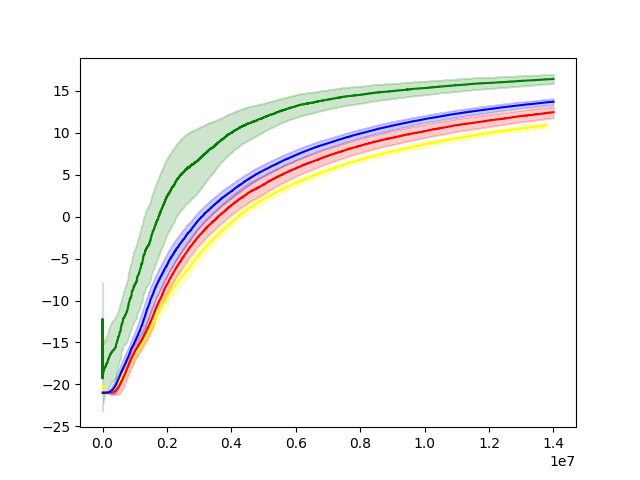}\vspace{-.3cm}
\\
(a)&(b)\\
 
  \end{tabular}
  \caption{\label{fig:mean} A side by side comparison between the mean reward over the entire training process and the current reward of the training for the various TL methods on the game Pong. The x-axis is the number of training steps, and the y-axis are the mean reward over the entire training until this point (a) or the current reward (b). The plots are averaged over three independent runs. The blue line is the baseline, the red is distillation, the yellow is continuous data transfer and the green is data transfer for pre training.}
\end{figure*}

Tab.~\ref{tab:bigsummary} shows the level of success reached by the various methods, in comparison to the baseline method. While the scoring is subjective, the table suggests that the first method of data transfer for the purpose of pretraining is the only method to consistently outperforms the baseline. Fig~\ref{fig:graphs} contains the full training logs. Note that using the Mean Reward Evaluation score, as done in~\cite{rusu2016progressive}, would indicate that our methods mostly outperform the baseline. However, this can happen with each of the level of success above as can be seen in Fig.~\ref{fig:mean}.

%since mean reward  is biased towards methods that are very stable in the beginning of the training process, or starts in a good way but not necessarily finish in the same way. We saw that even though our algorithms behave much better when evaluated using the mean reward, in most cases that did not indicate a good training process to the objective reward.

\begin{table*}[t]
\caption{The level of success (see text) reached by the various methods.}
\label{tab:bigsummary}
%\begin{center}
\centering
\begin{small}
\begin{sc}
\begin{tabular}{lccc}
\toprule
 & Data transfer& Continious data  & Distillation \\
Source $\Rightarrow$  Target & pretraining & transfer & \\
%Source $\Rightarrow$  Target			& Convolution Initialization & Co-Train & Pre-Train   \\
%Source $\Rightarrow$  Target			& Data transfer pretraining & Continious data transfer & Distillation   \\
\midrule
Breakout $\Rightarrow$ Pong			& *,2 & - & - \\
Pong $\Rightarrow$ Breakout			& *,2 & 2 & - \\
Tennis $\Rightarrow$ Pong			& 3   & - & - \\
Tennis $\Rightarrow$ Breakout		& -   & - & - \\
Breakout $\Rightarrow$ Tennis		& 1   & 1 & 1 \\
Pong $\Rightarrow$ Tennis			& 1   & 1 & 1 \\
Assault $\Rightarrow$ Demon-Attack	& 1 or 2   & - & - \\
Demon-Attack $\Rightarrow$ Assault	& 2   & - & 2 \\
\bottomrule
\end{tabular}
\end{sc}
\end{small}
%\end{center}
\vskip -0.1in
\end{table*}

\section{Results Analysis}

It is easy to identify potential shortcoming of visual analogies. First, the analogies are partial, e.g., the opponent in Pong does not exist in Breakout. Second, the dynamics are of a different nature. For examples, in the game Tennis, the game is a projection of 3D world to 2D, which makes the ball move in a slight curve and in different speeds depending on the ball location, unlike Pong or Breakout. There are additional differences such as the speed of the ball relative to the paddle, speed of the bullet relative to the spaceship and the targets, etc.

We try to augment the game's speed in order to tackle some of these issues. The speed issue arises when transferring between the games Tennis and Pong where the speed of the tennis player with respect to the ball is slower relative to the speed of the Pong paddle with respect to the ball. This leads to very different strategies and makes the Pong strategy much less relevant to the Tennis game. For example, a valid strategy in the Pong game is to wait until the last possible second and only than move to hit the ball with the side of the paddle. This strategy is impossible to apply in the Tennis game since the Pong agent will start moving too late in the Tennis game and the player will not be able to hit the ball on time. 

\begin{figure*}
%\centering
%\begin{minipage}[t]{.4\textwidth}
  \centering
  \includegraphics[width=.7\linewidth]{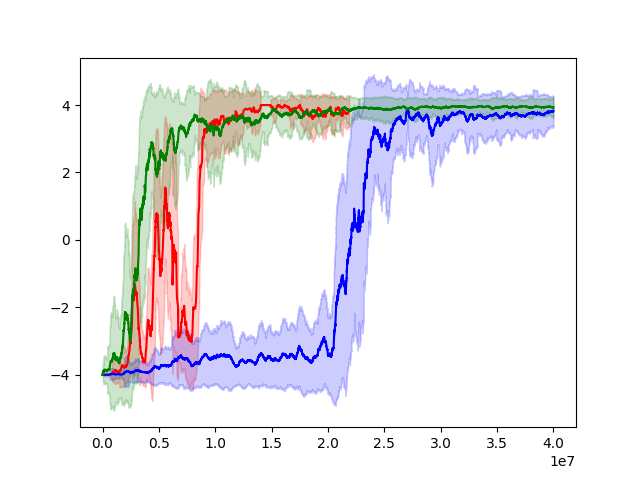}
  \caption{ A comparison of the training logs for the game Tennis pretrained with different speeds of Pong, Blue is the baseline (no transfer) green is Tennis pretrained with regular speed pong, and red is Tennis pretrained with slow pong (closer speed to tennis speed).}
  \label{fig:speed}
\end{figure*}

To overcome this problem and check how much it affects the final results, we designed a set of experiments where we used different relative speed for the pretrained Pong game in order to over come the speed difference and make the pong paddle speed closer to the tennis player speed. We created those games by tranforming the agent $\pi_s$ and inserting a NONE action after every few real actions played by the agent.
\[
\pi_{tranform}(s) =  \begin{cases}
\pi_s(s) \text{ If the index of the frame is divided by the <slowing factor>}\\
NONE \text{ Otherwise}\end{cases}
\]
This results in slowing down the relative speed between the paddle and the ball by the speed factor. Fig.~\ref{fig:speed} shows the results of the training of Tennis with the different Pong games variations for pretraining. As can be seen, changing the relative speeds before applying transfer using the data transfer by pretrainig did not improve the Tennis training time. This is probably due to the fact that there are many more factors that affects the transformation validity.

%In addition the slowing down is done with a certain granularity and not all speeds are possible. Specifically there is no way of making sure that the relative speed that we got is the "correct" one.

%Figure-<> shows training speed on the game Tennis with pre-trained Pong with different speeds of the paddle relative to the ball. it seems that changing only one such parameter (the relative speed of the paddle) doesn't affect the final results. This is reasonable because of two possible reasons, 1. in order for it to better affect the final results we need one of the speeds to exactly match the speed of the other game and it's hard to achieve with the current environment, and secondly there are a lot of other parameters that affect the gameplay and the agent are over fitting and some of them even affect the ball speed - such as acceleration and ball movement. For example in Tennis the ball moves in a curve but in pong the ball moves in a straight line which makes matching the speeds perfectly almost impossible.

%\noindent{\bf The Reward Function}
For some pairs of games the reward function is inherently different, which makes the agent optimize on the wrong objective. Such a problem arises in the pairs of Pong and Breakout and Tennis and Breakout, where there is a fundamental difference in the reward function. While in Breakout the objective is to break the most bricks before the player loses, in Pong and Tennis the objective is to make the other player lose. While both reward functions require the paddle of the player to hit the ball, they are very different in the way they want the ball to behave after being hit and they oppose each other in the amount of time they aim for the game to last -- in Breakout the more time you play the better the score is (since you have a better chance at breaking more bricks), while in Pong and Tennis the more time you play, the higher your chance of losing since every time you have to hit the ball is a time where your opponent succeeded in hitting the ball.

\begin{figure*}

\centering
  \includegraphics[width=.7\linewidth]{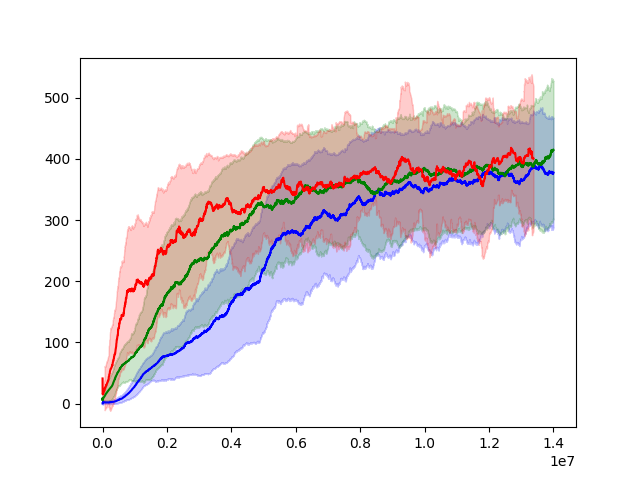}
  \caption{A comparison of the training logs for the game Breakout pretrained with different reward function for Pong, Blue is the baseline green is regular reward, and red is our improved reward.}
  \label{fig:reward}
%\end{minipage}
\end{figure*}

In order to make the pairs more similar we designed a different version of Pong where the reward function is the number of times the player hit the ball successfully.\\
\[R_{new~pong} = \begin{cases}
1 \text{ If the ball crosses the middle from bottom to top}\\
0 \text{ Otherwise}\end{cases}\]
Fig.~\ref{fig:reward} shows the difference in training time of games trained with the Data transfer for pretraining TL method where the pretraining on the source is done using our improved reward or the original reward. It can be seen from the graph that changing the reward function does improve the start of the training and reaches the asymptotic level of performance considerably faster than both transferring from the original game and the baseline. However, the asymptotic success is not significantly improved.

%, but as the training continues to reach the same asymptotic limit. %This is reasonable, since we assume that the reward function is indeed part of the problem, but it is only one part of the difference between the games, and this is probably the reason why the agent starts to overfit on the current game.

%\noindent{\bf The Solution}
%As can be seen from the above experiments, solving those problems by trying to overcome every small difference between the games directly is impossible - since there are many small differences and we do not have access to changing most of them - for example in the game Pong the ball moves in a constant speed but in Breakout the ball accelerates, we have no way of accelerating the ball in Pong without changing the state transformation function which is impossible.\\
%Our possible solution is to approach all of those problems as one by understanding its source. We claim that those problems are actually one problem which is known to happen in  Reinforcement Learning - Overfitting. Since in Reinforcement Learning we usually do not have a test set - and we only train on one environment, agents have the tendency to overfit to this  specific environment, and this could be a possible explanation as to why the the transfer learning methods do not work well in our settings, even though the transformation function seems to work. This also explains why the environments need to be almost the same after the transformation in order for the transfer learning to work.\\

\begin{figure*}
\centering
\begin{minipage}[t]{.7\textwidth}
  \centering
  \includegraphics[width=1\linewidth]{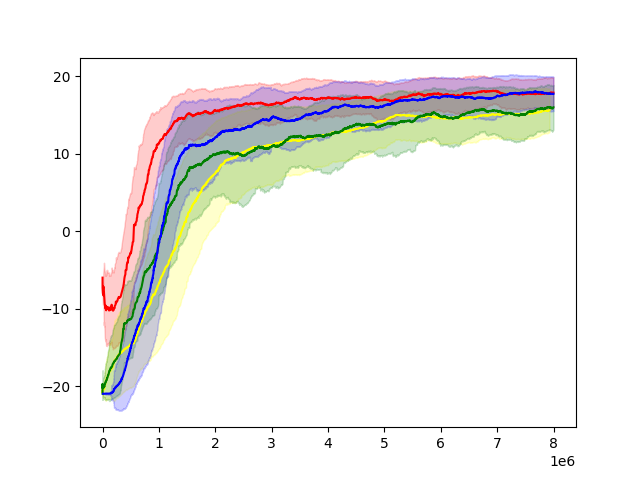}
  \caption{A comparison of the training logs for the game Pong pretrained with breakout or breakout and pong. Blue is the baseline, green is Pong pretrained with breakout, yellow is Pong pretrained with Tennis and red is Pong pretrained with breakout and Tennis.}
  \label{fig:multi}
\end{minipage}\qquad
\end{figure*}

A more general view of the failure to transfer would conclude that the policy $\pi_s$ overfits the source game and is not able to generalize (RL methods are known to be prone to overfitting). Instead of trying to overcome this by making the source game more similar to the target game, we next try to train a model to fit two source games. Namely, we train a Pong agent in the following way: first we trained a combined model for transfered frames from Tennis to pong and transfered frames from Breakout to Pong. Once the model performed well on both the transfered games, we trained this model on the original Pong frames. Fig.~\ref{fig:multi} shows that this leads to reaching the asymptotic performance faster than pretraining with any of the two games separately.

%NOTE THAT THE FIGURE IS MISSING THE ONLY TENNIS BASELINE. 
%TRY TO HAVE THE ORDER OF THE FIGURES IN THE WRITING MATCH THE ORDER OF APPEARANCE.

%Fig.~\ref{fig:graphs} shows the training graphs of the transfer learning methods. Each point on the graphs is an average of samples of the model from the last 100K training states. The plotted results are the average of three independent runs.

\clearpage

\bibliography{gans,rl,tmann,7057-65934-BIB,7057-65876-BIB}
\bibliographystyle{iclr2018_workshop}

%\section{Additional Training Graphs}
%\label{app:b}

%\noindent{\bf Fig.~\ref{fig:speed}} shows the training logs for the game Tennis pretrained with different speeds of Pong. As can be seen, trying to match the speed of Pong with the speed of Tennis does not improve the training process of Tennis in an apparent way.

%\noindent{\bf Fig.~\ref{fig:reward}} shows the training logs for the game Breakout pretrained with different reward functions for Pong. As can be seen, matching the reward of the games improves the reward in the first epochs of the training, but as the training continues, the baseline catches up.

%\noindent{\bf Fig.~\ref{fig:multi}} shows the training logs for the game Pong with multiple games in the pretraninig phase. As can be seen, training on multiple games in the pretrain phase of the Data Tranfer by Pretraining method improves the training speed.

%\end{minipage}\qquad
%\begin{minipage}[t]{.4\textwidth}

\end{document}